\PassOptionsToPackage{dvipsnames}{xcolor}
\documentclass{article} 
\usepackage[preprint]{colm2025_conference}

\usepackage{microtype}
\usepackage{hyperref}
\usepackage{url}
\usepackage{booktabs}
\usepackage{graphicx}
\usepackage{subcaption}
\usepackage{lineno}
\usepackage{soul}
\usepackage{amsmath}
\usepackage{amssymb}
\usepackage{colortbl}
\usepackage{nicefrac}       
\usepackage{diagbox}
\usepackage{xcolor}
\usepackage{multirow}
\usepackage[normalem]{ulem}
\useunder{\uline}{\ul}{}
\usepackage[all=normal, mathspacing=tight,mathdisplays=normal, floats=tight, paragraphs=tight, lists=normal]{savetrees}
\definecolor{darkblue}{rgb}{0, 0, 0.5}
\hypersetup{colorlinks=true, citecolor=darkblue, linkcolor=darkblue, urlcolor=darkblue}

\newcommand{\quotes}[1]{``#1''}

\title{Exploiting Mixture-of-Experts Redundancy Unlocks Multimodal Generative Abilities}

\author{\small{Raman Dutt${}^{\spadesuit}$, Harleen Hanspal${}^{\diamond}$, Guoxuan Xia${}^{\diamond}$, Petru-Daniel Tudosiu${}^{\clubsuit}$, Alexander Black${}^{\dagger}$,}\\ \small{\textbf{Yongxin Yang${}^\blacklozenge$, Steven McDonagh${}^{\spadesuit}$, Sarah Parisot}${}^{\ddagger}$}
\\\\
$^{\spadesuit}$ The University of Edinburgh \\
$^{\diamond}$ Imperial College, London \\
$^{\clubsuit}$ Leonardo.AI \\
$^{\dagger}$ University of Surrey \\
$^{\blacklozenge}$ Queen Mary University of London \\
$^{\ddagger}$Microsoft Research, Cambridge\\\\
\small{\texttt{raman.dutt@ed.ac.uk, h.hanspal21@imperial.ac.uk, g.xia21@imperial.ac.uk}}\\
\small{
\texttt{daniel.tudosiu@leonardo.ai, alexander@black.com}
}\\
\small{
\texttt{yongxin.yang@qmul.ac.uk, s.mcdonagh@ed.ac.uk, sarahparisot@microsoft.com}
}
}

\begin{document}

\ifcolmsubmission
\linenumbers
\fi

\maketitle


\begin{abstract}
In this work, we undertake the challenge of augmenting the existing generative capabilities of pre-trained \textit{text-only} large language models (LLMs) with multi-modal generation capability while satisfying two core constraints: \textbf{(C1)} 
preservation of original language generative capabilities with negligible performance degradation, and \textbf{(C2)} adhering to a small parameter budget to learn the new modality, ensuring scalability and efficiency. In contrast to current approaches that add dedicated modules, thereby significantly increasing the parameter count, we propose a method that leverages the underutilized capacity inherent in deep models. Specifically, we exploit the parameter redundancy within Mixture-of-Experts (MoEs) as a source of additional capacity for learning a new modality, enabling better parameter efficiency (C1). Moreover, we preserve the original language generation capabilities by applying low-rank adaptation exclusively to the tokens of the new modality (C2). Furthermore, we introduce a novel parameter initialization scheme based on the Gromov-Wasserstein distance to improve convergence and training stability. Through an extensive analysis of the routing mechanism, we uncover the emergence of modality-specific pathways and decreased redundancy within the experts that can efficiently unlock multi-modal generative capabilities. Overall, our method can be seamlessly applied to a wide range of contemporary LLMs, providing a new pathway for transitioning from uni-modal to multi-modal architectures.

\end{abstract}

\section{Introduction}

Autoregressive modelling via next-token prediction has demonstrated a remarkable capacity for semantic processing and generating language at scale~\citep{radford2019language}. Large Language Models (LLMs), trained exclusively on textual data, have achieved groundbreaking milestones, such as outperforming clinicians~\citep{kim2024large} and earning silver-medal performance at international olympiads~\citep{trinh2024solving}. Building on these successes, the next logical step is to transcend the textual modality and tackle diverse data forms. Although considerable research has focused on adapting LLMs for visual understanding~\citep{liu2023llava,zhao2024tuning,team2024chameleon}, extending these models to incorporate image generation remains an emerging and challenging frontier~\citep{pmlr-v239-zhang23a,sun2024autoregressive,ashutosh2025llms,sun2024autoregressive,shi2024llamafusion}.

While the sequential structure of language naturally supports next-token prediction, diffusion models have emerged as the prevailing paradigm for image generation \citep{song2019generative,ho2020denoising,dhariwal2021diffusion,rombach2022high}, concurrently generating image content through iterative denoising. Nevertheless, recent work on multi-modal generative models has demonstrated the feasibility of employing next-token prediction for image generation \citep{sun2024autoregressive,dai2023emu,jin2024unified}. One popular strategy involves fine-tuning pre-trained LLMs on multi-modal (text-image) data \citep{ge2024making,he2024mars}, thereby leveraging their complex language understanding for text-to-image tasks while mitigating the high costs associated with training on large-scale language corpora. 



Although effective, fine-tuning LLMs on multi-modal data frequently degrades their original text performance \citep{he2024mars}. To mitigate this degradation, researchers commonly augment training datasets with additional text-specific data to preserve or even enhance the LLM’s original text-understanding and generation capabilities \citep{jin2024unified,team2024chameleon}. This approach can be computationally expensive and may undermine the benefits of leveraging pre-trained LLMs. Consequently, researchers have explored an alternative strategy: integrating new, learnable modality-specific weights into the frozen LLM architecture \citep{he2024mars,ge2024making}. While this solution is appealing because it preserves the original LLM’s capabilities and enables strong image generation capabilities, it suffers from significant parameter \textit{inefficiency} and poor scalability. For instance, the \textit{SemVIE} module introduced in \citet{he2024mars} more than doubles the model’s parameter count to accommodate a new modality, rendering the approach computationally expensive and impractical for scaling to multiple input modalities or larger models. This necessitates a solution that can integrate modality-specific parameters without incurring prohibitive computational costs.

\begin{figure}[bht]
\begin{center}
\includegraphics[width=0.8\columnwidth]{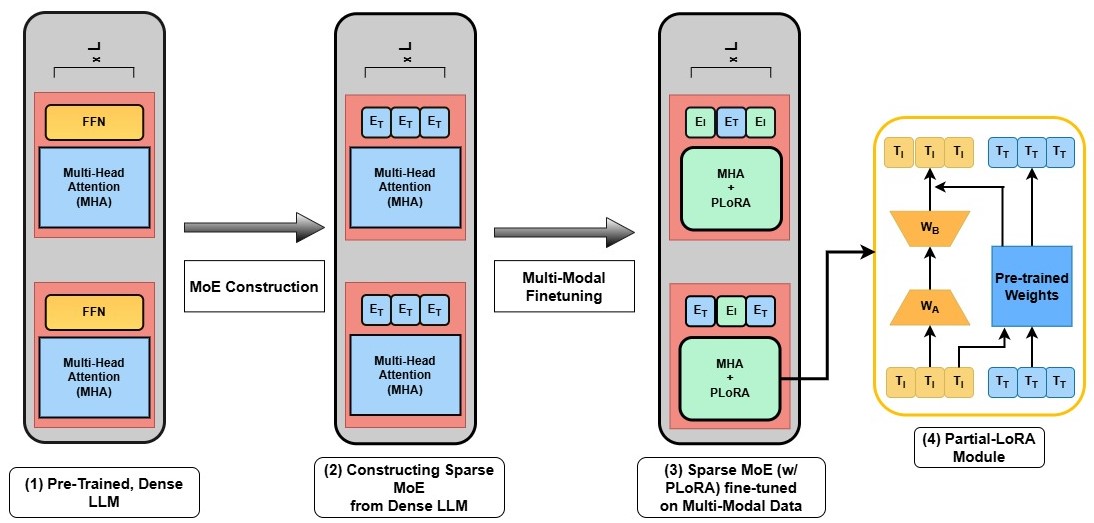}
\end{center}
\vspace{-3mm}
\caption{Overall schematic of the proposed framework. \textbf{(1)} Dense pre-trained LLM is converted to its MoE variant. \textbf{(2)} Each expert in LLM-MoE is still a \emph{text-expert} due to text pre-training. \textbf{(3)} The MHA block in the LLM is then modified with the PLoRA module and fine-tuned on multi-modal data. During fine-tuning, the routers learn to assign dedicated experts to image and text modalities. 
\textbf{(4)} We illustrate the PLoRA module, which applies low-rank adaptation exclusively to the image tokens in an input sequence containing both image (yellow) and text tokens (blue).}\label{fig:schematic}
\vspace{-1mm}
\end{figure}

In this work, we leverage the inherent redundancy in large language models, where many layers perform equivalent operations, to unlock latent capacity for learning new modalities. While previous research has sought to eliminate redundant parameters for efficiency gains \citep{ma2023llm,men2024shortgpt,ashkboos2024slicegpt}, we posit that this redundancy can be repurposed to facilitate multi-modal learning. 
Our framework begins by converting a pre-trained, dense LLM (LLM-Dense) into its Mixture-of-Experts variant (LLM-MoE) \citep{jacobs1991adaptive, shazeer2017}. The MoE architecture is particularly appealing due to its demonstrated high expert redundancy \citep{chen2022task,sarkar2024revisiting,li2024merge,moe_pruner,chen2023sparsemoenewdropout} and its potential ability to construct dedicated pathways through expert routing for different modalities. Additionally, we introduce two classes of new parameters into the model: \textbf{(1)} low-rank adapters within the transformer decoder blocks, and \textbf{(2)} encoding and decoding parameters in the embedding and head layers. For the former, we adopt a \textit{partial} low-rank adaptation (PLoRA) scheme that updates adapters exclusively with tokens from the new modality \citep{dong2024internlm}, thereby preserving the original language generation performance and eliminating the need for additional text-only fine-tuning. For the latter, we propose a novel parameter initialization method based on the Gromov-Wasserstein (GW) distance \citep{memoli2011gromov}, which improves cross-modality alignment and enhances stability and convergence during multi-modal fine-tuning.
Collectively, these strategies enable us to extend pre-trained uni-modal LLMs to multi-modal generation in a data and parameter-efficient manner without compromising their original abilities. Our experiments show that our approach delivers strong and competitive image generation performance using very modest data (7.5 million training samples), parameter and compute budget.
An overview of our framework is presented in Figure \ref{fig:schematic}.

\section{Related Work}
This work builds upon recent advances in token-based multi-modal learning and mixture-of-experts frameworks. In the context of token-based autoregressive approaches for text-to-image generation, prior work can be broadly categorized into two groups. The first comprises methods that natively integrate both text and image modalities into their pre-training datasets \citep{team2024chameleon,chern2024anole,ge2024making}. The second group focuses on evolving the vision generation capabilities of an existing pre-trained language-only model \citep{dong2023dreamllm, zhan2024anygpt, liu2025worldmodelmillionlengthvideo}. To leverage the rich, semantic knowledge inherent in pre-trained LLMs and to incrementally extend them to new modalities in a data- and parameter-efficient manner, we adopt the latter approach and extend a pre-trained LLM to process the vision modality.

One of the primary challenges in extending a pre-trained LLM to new modalities is catastrophic forgetting of the original text modality~\citep{he2024mars}. To mitigate this, existing works employ careful data and loss balancing to preserve the original performance while incorporating new modalities \citep{liu2025worldmodelmillionlengthvideo}. Alternatively, \citep{he2024mars} alleviates the need for such delicate balancing by adding an expert exclusively for the vision modality. In general, the use of (mixture of) experts has enabled faster model training and inference, as well as a more modular and interpretable expansion of the model input domain. However, the additional expert in \citep{he2024mars} doubles the model parameter count, making it expensive to implement and less scalable to integrate further modalities. Consequently, our work aims to retain the benefits of modality-specific experts while minimizing the parameter overhead.

Several prior works have explored constructing Mixture-of-Experts (MoE) models from existing dense LLMs while preserving the original model parameter count. These methods commonly partition the parameters of the Feed Forward Networks (FFNs) in the LLM, differing primarily in the specific techniques used for expert construction and token routing. For example, \cite{zhang2021moefication} exploited highly sparse and correlated neuron activations to partition a FFN into distinct experts. While their approach explicitly creates experts using algorithms such as K-Means clustering and co-activation graph splits, we find that a random, non-overlapping partitioning of FFN units is sufficient to facilitate the learning of experts from the FFN layers. Similarly, token routing can be implemented either by selecting the top-K experts for each token, as is typically done \citep{tokenchoice}, or by choosing the top-K tokens for each expert, as proposed in \citet{zhou2022mixture}. Given that our objective is autoregressive generation, we adopt the former routing approach. 
Similar to our approach, \cite{zhu2024llama} introduced LLaMA-MoE by partitioning the FFN layer of a LLaMA-2-7B model \citep{touvron2023llama} into multiple experts. They further applied continual pretraining to optimize the modified model, achieving state-of-the-art performance among open-source models that convert dense pre-trained LLMs to MoEs. While LLaMA-MoE is tailored exclusively for natural language processing, our work extends the MoE framework to both language and image domains, addressing multimodality-specific challenges such as vocabulary expansion and initialization. 
Our approach also differs from native multi-modal MoE methods~\citep{unimoe, aria}, which pre-train modality-specific experts separately on cross-modality data and then jointly fine-tune them using LoRA \citep{hu2022lora, dutt2024parameterefficient}. In contrast, we introduce the visual modality solely during the low-rank fine-tuning of a pre-trained text-only LLM. Consequently, our multi-modal MoE requires significantly less data and training compared to native multi-modal MoE models.

\section{Preliminary}

\subsection{Mixture of Experts}
A typical Mixture of Experts (MoE) module is similar to a standard transformer module~\citep{vaswani2017attention} where the FFN modules are replaced with MoE Layers consisting of $N$ expert networks and a gating network $G$. In sparse MoEs, the number of active experts, $K$, is a fixed value significantly smaller than the total number of experts, $N$. The gating module selects the Top-$K$ experts for each input token. Formally, for a given input $x$, let $E_i(x)$ represent the output of the $i$-th expert; the output of the MoE layer is computed as the sum of the outputs from the $K$ selected experts.

\begin{equation}
    y = \sum_{i=1}^NG(x)_i * E_i(x), \quad  G(x)_i = 
\begin{cases} 
s_{i,t}, & \text{if } s_{i,t} \in \text{Top-}K(\{s_{j,t} \mid 1 \leq j \leq N\}, K), \\
0, & \text{otherwise}.
\end{cases}
\end{equation}


where $s_{i,t}$ denotes the expert score for the $t$-th token and is obtained by applying Softmax to the output of the gating network $G$. Top-$K$ denotes the $K$ highest scores for the $t$-th token.



To address the issue of load balancing, i.e., the over- and under-utilization of experts during training, the Noisy Top-$K$ gating mechanism~\citep{shazeer2017} has become a widely adopted approach for MoE models. This mechanism adds tunable Gaussian noise to the gating Softmax's inputs to smoothen the expert scores and promote effective load balancing. This noise is set to zero at the start of training to ensure an equal distribution of load across all experts and gets tuned during training to do sparse expert routing.

\subsection{Discrete Image Tokenization for Autoregressive Modeling}


LLMs are trained using the next-token prediction loss on discrete tokens. Accommodating a new modality in the same architecture first requires discrete tokenization of the new modality using modality-specific encoders. For images, this tokenization is typically performed using a Vector-Quantized Variational AutoEncoder (VQ-VAE) \citep{van2017neural}. A VQ-VAE transforms the input image pixels $x$ into a corresponding feature map $f$ and assigns each vector $f^{(i,j)}$ in the feature map to the index $q^{(i,j)}$ of its closest codebook vector $z^{(i,j)}$. During decoding, the indices $q^{(i,j)}$ are mapped back to their respective codebook vectors $z^{(i,j)}$, which are then reconstructed into the image pixels $\hat{x}$ by the decoder. 
For our work, we employ the image tokenizer in \citet{sun2024autoregressive}, which adopts an encoder-quantizer-decoder architecture similar to that of \citet{esser2021taming}. This tokenizer has a codebook size of 16384, a downsampling ratio of eight, and achieves a reconstruction quality (r-FID) of 2.19 for 256$\times$256 images using ImageNet~\citep{deng2009imagenet}. Further details about this process are provided in Appendix \ref{sec:mm_generation_appendix}.

\section{Methodology}

\subsection{From Dense LLM to Mixture-of-Experts} \label{sec:llm_moe}

We select the Mixture-of-Expert architecture due to two main reasons: \textbf{(1)} it can enable the creation of modality-specific pathways by assigning different experts to distinct modalities, and \textbf{(2)} to leverage the inherent expert redundancy as latent capacity to learn the new modality. Therefore, the first step in our framework is to construct a sparse MoE from a dense LLM (LLM-Dense $\rightarrow$ LLM-MoE). To achieve this, we adopt the framework presented in \citet{zhu2024llama} and construct MoE experts by splitting the Feed Forward Layers (FFN) of the dense LLM. Amongst several possible strategies for creating these splits, \citet{zhu2024llama} found \textit{neuron-independence} as the most effective where FFN neurons are randomly partitioned into equal-sized groups to form the experts such that no two neurons belong to the same expert. \\
This procedure is independent of the LLM architecture and can be easily applied to obtain the MoE variant of any given LLM. In this work, we use the LLaMA architecture \citep{touvron2023llama}, following \cite{zhu2024llama}.

\subsection{Preserving Language Abilities while Enabling Multi-Modal Generation} \label{sec:plora}
Extending unimodal LLMs to multiple modalities has been previously achieved using low-rank adapters \citep{hu2022lora} and multi-modal fine-tuning \citep{su2023pandagpt,zhang2023llama}. These solutions update model weights and low-rank matrices with both image and text tokens, which can degrade the model's original text generation capabilities (see Tab. \ref{tab:nlp_results}). We hypothesize that a pre-trained text LLM requires only to adapt to the new modality (images), leaving text unchanged. This can be accomplished by introducing low-rank adapters solely for image tokens, an approach termed Partial LoRA (PLoRA; \citep{dong2024internlm}). We anticipate that modality specific adaptation will diversify the routing process, directing image tokens to experts that were previously less frequently selected by text tokens (redundant experts). We further antiticpate this to enable formation of image-specific pathways while preserving language abilities. We introduce the low-rank adapters in the \textit{query}, \textit{key}, \textit{value}, and \textit{out} projection layers. During training, we set them as trainable along with the MoE router and MoE experts.
\\
Formally, for each linear layer \( L_0 \) in the LLM, we represent its weight matrix as \( W_0 \in \mathbb{R}^{C_{\text{out}} \times C_{\text{in}}} \) and its bias as \( B_0 \in \mathbb{R}^{C_{\text{out}}} \), where \( C_{\text{in}} \) and \( C_{\text{out}} \) denote input and output dimensions, respectively. Similar to LoRA, PLoRA comprises two low-rank matrices, \( W_A \in \mathbb{R}^{C_r \times C_{\text{in}}} \) and \( W_B \in \mathbb{R}^{C_{\text{out}} \times C_r} \). For a given input \( x = [x_v, x_t] \), the text tokens ($x_t$) are processed with the original pre-trained weights, $W_o$, while the image tokens ($x_v$) are passed through both the original pre-trained weights $W_o$ and the trainable low-rank weights $W_BW_A$. We depict this procedure in Fig.~\ref{fig:schematic} and formalize the low-rank representations accordingly in Eq.~\ref{eqn:plora} as

\begin{equation}
\begin{aligned}
    \hat{x}_v = (W_o + W_BW_A)x_v + B_o, \\
    \hat{x}_t = W_ox_t + B_o, \\
    \hat{x} = [\hat{x}_v, \hat{x}_t].
\end{aligned}
\label{eqn:plora}
\end{equation}

\subsection{Parameter Initialization with Gromov-Wasserstein Distance} \label{sec:gw_init}
Adapting a pre-trained language model to incorporate a new modality necessitates adding parameters in the embedding and head layers to encode and decode the new modality tokens as new vocabulary. Previous studies have often employed simplistic initialization strategies, such as random initialization \citep{ge2024making} or using the mean of existing parameters \citep{he2024mars}. However, we contend that these approaches are suboptimal and do little to promote cross-modality alignment. We hypothesize that the new parameter sets should be initialized from a distribution that closely mirrors the existing text embeddings, facilitating better cross-modal alignment and successfully leveraging pre-trained weights.
For this task, we propose a novel parameter initialization scheme that leverages the Gromov-Wasserstein (GW) distance \citep{memoli2011gromov}. Our objective is to initialize the new parameters (image embeddings) by aligning them with the distributional properties of existing parameters (text embeddings) to ensure compatibility during fine-tuning.


Let the pre-existing embedding spaces of text and image tokens be denoted as \mbox{$E_t \in \mathbb{R}^{|V_t| \times d}$} and \mbox{$E_i \in \mathbb{R}^{|V_i| \times d}$}, respectively, where $d$ is the embedding dimension, and $|V_t|$ and $|V_i|$ are the sizes of the text and image vocabularies. The GW distance measures how well the pairwise distance distributions between the two sets can be aligned. This objective seeks to find the optimal coupling $\gamma^*$ that minimizes the distance between the two embedding spaces. Specifically, the optimal coupling is obtained by solving an optimization problem that minimizes the difference between the pairwise distance matrices of the text and image embeddings. Once $\gamma^*$ is obtained, we initialize the new image embeddings by solving equation \ref{eqn:gw_init}.
\begin{equation} \label{eqn:gw_init}
    E_i = 
    \underset{E_i'}{\mathrm{argmin}}
    \sum_{x \in V_t} \sum_{y \in V_i} \gamma^*(x, y) \| E_t(x) - E_i'(y) \|^2.
\end{equation}

This initialization strategy aligns the pairwise distance distributions of the new image embeddings with those of the pre-trained text embeddings, effectively bridging the gap between the two modalities. Consequently, the geometric relationships inherent to the text embedding space are transferred to the image embedding space, enabling better cross-modality alignment. Moreover, this approach allows the MoE from Section~\ref{sec:llm_moe} to capture shared aspects of both modalities, while PLoRA focuses on learning the unique characteristics of the new modality. This, in turn, facilitates the formation of \emph{modality-specific experts}.
In Fig. \ref{fig:loss_conv}, we illustrate that GW initialization facilitates faster convergence and stability during the fine-tuning process as compared to other initialization schemes.


\subsection{Continual Pre-Training with Multi-Modal Data} We perform fine-tuning on multi-modal data after constructing the LLM-MoE (Sec. \ref{sec:llm_moe}), introducing low-rank adapters (PLoRA, Sec. \ref{sec:plora}), and introducing and initializing new parameters in the embedding and head layers using the Gromov-Wasserstein initialization (Sec. \ref{sec:gw_init}). We divide our training process into two stages: \textbf{Low-Res Training} and \textbf{High-Res Training}. The former stage enables initial text-to-image alignment while the later stage helps improve aesthetics using high-resolution data. 
We purposely used high-quality, photorealistic 4K images as improved data quality has been found to improve the generation quality \citep{chen2025pixart}. 
Furthermore, we employed Share-Captioner \citep{chen2025sharegpt4v}, which generates coherent and detailed captions for each image, increasing the average caption length to 180 words. For \textbf{Low-Res} training, we employed a subset of 4M samples at a 256$\times$256 resolution, training the model for five epochs. In the \textbf{High-Res} stage, we employed another subset of 3.5M samples at 512$\times$512 resolution, continuing the training for five additional epochs. During training, we set the new encoding and decoding parameters in embedding and head layers, PLoRA parameters, MoE router layers, and MoE expert layers as trainable.

\section{Experiments}

\textbf{Experimental Settings: }We base our experiments on the LLaMA-MoE (4/16) model \citep{zhu2024llama}, an MoE variant derived from the dense LLaMA-2-7B model \citep{touvron2023llama}. This model comprises 16 experts per layer, activates the top four experts for each input token, and activates approximately 3.5B parameters. We introduce low-rank adapters with a rank of 64 and apply rank stabilization \citep{kalajdzievski2023rankstabilizationscalingfactor}. For optimization, we employ a learning rate of 2e-4, which decays to 2e-5 using cosine scheduling with 1,000 warmup steps. We train for five epochs in both \textbf{Low-Res} and \textbf{High-Res} training stages. To improve training efficiency, we leverage DeepSpeed ZeRO-3 \citep{rajbhandari2022deepspeed}, FlashAttention v2 \citep{dao2023flashattention}, and Liger Kernels \citep{liger_kernels}. All experiments are conducted on eight NVIDIA L40 GPUs 
with a per-device batch size of 16 and a gradient accumulation of four.

\subsection{Preserving Original Language Abilities During Multi-Modal Fine-Tuning}

\begin{table}[h]
\resizebox{\textwidth}{!}{%
\begin{tabular}{@{}ccccccccccc@{}}
\toprule
\diagbox{\textbf{Model}}{\textbf{Task}} & \textbf{Winogrande} & \textbf{NQ\_Open} & \textbf{Hellaswag} & \textbf{MMLU} & \textbf{Lambada} & \textbf{Arc-e} & \textbf{Arc-c} & \textbf{Piqa} & \textbf{SciQ} & \textbf{Average} \\ \midrule
\textbf{LLaMA-MoE}                                                   & 65.60               & 20.30             & 73.48              & 39.91         & 69.50            & 65.82          & 44.20          & 77.90         & 87.60    & 60.47     \\ \\
\begin{tabular}[c]{@{}l@{}}\textbf{LLaMA-MoE}\\ + \textbf{LoRA}\end{tabular}  &   43.27                  &      10.18             &       59.44             &     22.08          &       52.36           &     54.27           &      32.10          &     67.38          &      65.04   &  45.12 \color{red}(-15.35 $\downarrow$)    \\ \\
\rowcolor{gray!15} \begin{tabular}[c]{@{}l@{}}\textbf{LLaMA-MoE}\\ + \textbf{PLoRA (Ours)}\end{tabular} & 65.35        & 20.06      & 73.30      & 39.73  & 69.45     & 65.60   & 44.12   & 77.85  & 87.53 & 60.33 \color{teal}($\approx$-0.14)  \\ \bottomrule
\end{tabular}
}
\caption{Performance comparison on text benchmarks for the original LLaMA-MoE, LLaMA-MoE fine-tuned using LoRA, and LLaMA-MoE fine-tuned using PLoRA. While naive LoRA significantly degrades text performance, PLoRA can be seen to preserve the original text capabilities (\textcolor{red}{-15.35\%} and \textcolor{teal}{-0.14\%} respectively as compared to the original average performance).}
\label{tab:nlp_results}
\end{table}

Tab. \ref{tab:nlp_results} presents empirical results demonstrating how modality specific routing using LLaMA-MoE + PLoRA (Sec. \ref{sec:plora}) effectively preserves the original language generation capabilities of the LLM. Specifically, 
we compare the performance using our approach, with \textbf{(1)} the original model's reported performance (LLaMA-MoE) and \textbf{(2)} applying low-rank adaptation to both image and text tokens (LLaMA-MoE + LoRA) across the standard evaluation benchmarks. \\The results show that simply applying low-rank adaptation across both modalities significantly harms the original capabilities of the LLM, indicated by an average performance drop of \textcolor{red}{15.35\%} across all tasks. We anticipate a similar (or greater) performance drop in the case of full fine-tuning of the model. On the other hand, our modality specific routing using the PLoRA approach results in a negligible average performance degradation of \textcolor{teal}{0.14\%}, suggesting that redundant parameters were repurposed.

\subsection{Image Generation Quality}

\begin{figure}[ht]
\begin{center}
\includegraphics[width=0.90\columnwidth]{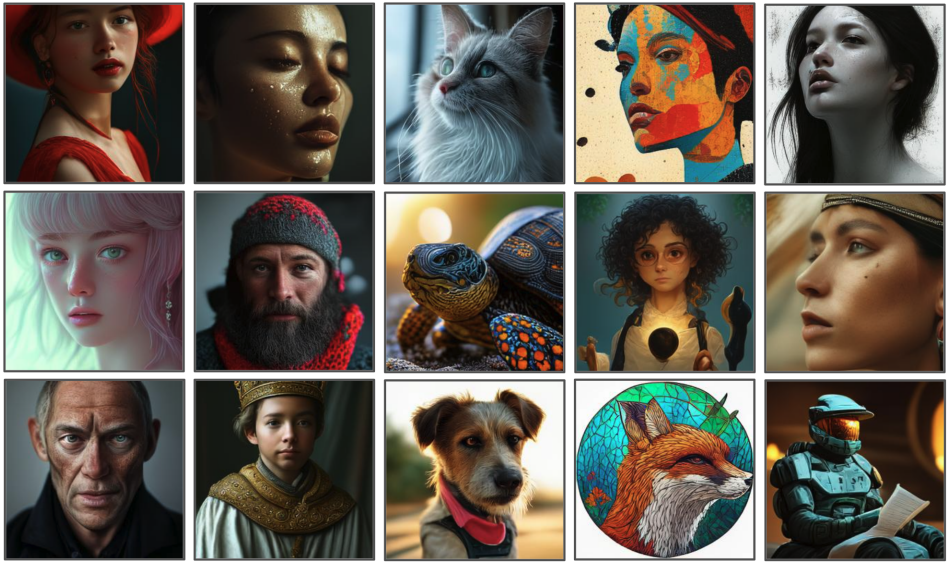}
\end{center}
\caption{Example generated samples using our approach. The images exhibit high fidelity and maintain strong textual coherence. See Appendix \ref{sec:text_coherence}, \ref{sec:more_examples}, and \ref{sec:gen_prompts} for examples showing strong textual coherence, more generated samples, and the associated prompts, respectively.}
\label{fig:generated_samples}
\end{figure}

\begin{figure}[htb]
    \centering
    \begin{minipage}{0.48\textwidth}
        \centering
        \resizebox{\textwidth}{!}{
        \begin{tabular}{@{}lccc@{}}
            \toprule
            \diagbox{\textbf{Metric}}{\textbf{Dataset}}                                                                       & \textbf{MSCOCO} & \textbf{CUB} & \textbf{Oxford} \\ \midrule
            \textbf{\begin{tabular}[c]{@{}l@{}}FID $\downarrow$\\ (LLaMA-MoE + \textcolor{BurntOrange}{LoRA})\end{tabular}}              & 12.58           & 7.13         & 8.96            \\
            \rowcolor{gray!15}\textbf{\begin{tabular}[c]{@{}l@{}}FID $\downarrow$\\ (LLaMA-MoE + {\textcolor{ForestGreen}{PLoRA}})\end{tabular}}             & 12.57           & 7.15         & 8.97            \\
            \textbf{}                                                                              &                 &              &                 \\
            \textbf{\begin{tabular}[c]{@{}l@{}}IS $\uparrow$\\ (LLaMA-MoE + \textcolor{BurntOrange}{LoRA})\end{tabular}}  & 28.20           & 6.36         & 4.28            \\
            \rowcolor{gray!15}\textbf{\begin{tabular}[c]{@{}l@{}}IS $\uparrow$\\ (LLaMA-MoE + {\textcolor{ForestGreen}{PLoRA}})\end{tabular}} & 28.20           & 6.35         & 4.26            \\ \bottomrule
            \end{tabular}
            }
            \captionof{table}{FID score (FID) and Inception Score (IS) for LLaMA-MoE fine-tuned using \textcolor{BurntOrange}{LoRA} and \textcolor{ForestGreen}{PLoRA} on MSCOCO, CUB and Oxford datasets. 
            }
            \label{tab:img_gen_results}
    \end{minipage}
    \hfill
    \begin{minipage}{0.48\textwidth}
        \centering
        \includegraphics[width=\columnwidth]{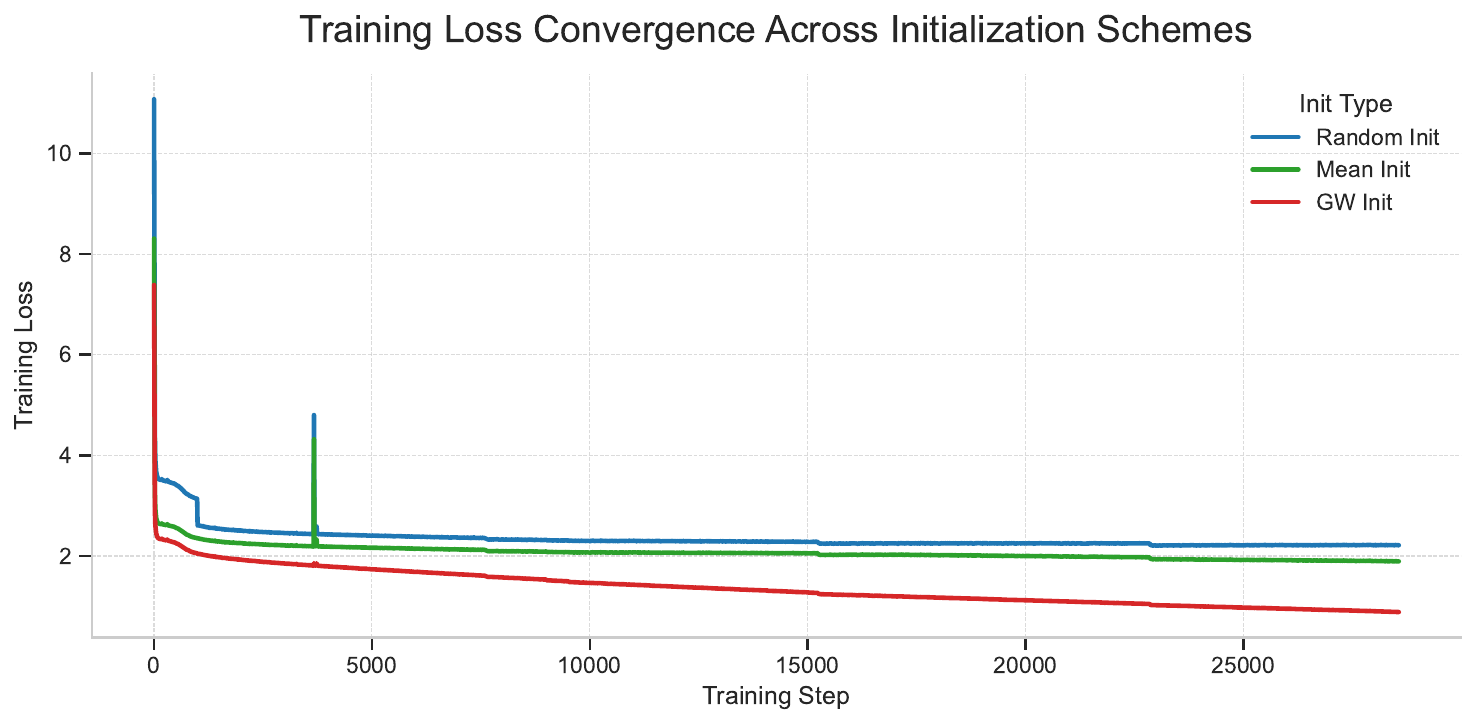}
        \captionof{figure}{Comparison of training loss convergence behaviour across different parameter initialization schemes. 
        }
        \label{fig:loss_conv}
    \end{minipage}
\end{figure}

We evaluate the image generation quality on MSCOCO (2017 Validation) \citep{lin2014microsoft}, CUB \citep{wah2011caltech,reed2016learning} and Oxford-102 \citep{nilsback2008automated} datasets using the Fréchet Inception Distance (FID) \citep{heusel2017gans} and Inception Score \citep{salimans2016improved}. Qualitative and quantitative results are presented in Fig. \ref{fig:generated_samples} and Tab. \ref{tab:img_gen_results} respectively. 
Specifically, we compare the performance of fine-tuning with LoRA (LLaMA-MoE + LoRA) and PLoRA (LLaMA-MoE + PLoRA). The results demonstrate that PLoRA does not compromise image generation ability, with both approaches achieving similar performance. For context, we achieve FID on MS-COCO on par with a latent diffusion model \citep{rombach2022high}, trained with $400$ million images (v.s. $7.5$ million ours). We additionally anticipate a distribution gap between our high-resolution aesthetic images and the MS-COCO dataset.  

\subsection{Analysis of Expert Routing}
In this section, we provide an in-depth analysis of the latent modality-specific routing mechanism. In Section \ref{sec:redundancy}, we compare the average redundancy among experts, before and after multi-modal fine-tuning, highlighting their role in facilitating multi-modal generation. In Section \ref{sec:modality_specific_agnostic}, we visualize the routing preferences of tokens from different modalities, which ultimately give rise to both \textit{modality-specific} and \textit{modality-agnostic} experts.

\subsubsection{Multi-Modal Fine-Tuning Reduces Expert Redundancy} \label{sec:redundancy}

\begin{figure}[ht]
\begin{center}
\includegraphics[width=0.95\columnwidth]{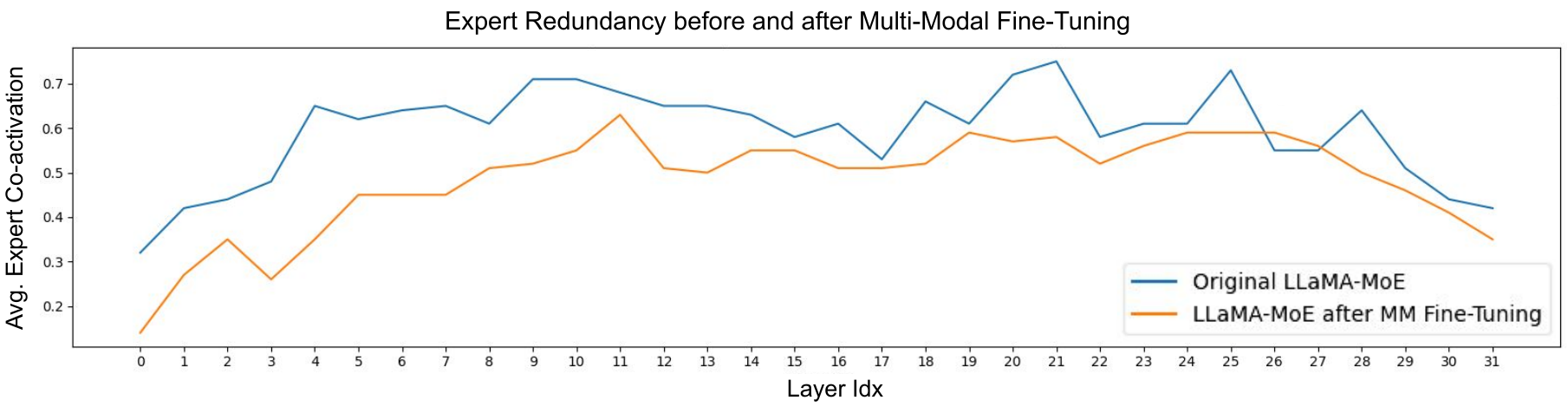}
\end{center}
\vspace{-13pt}
\caption{Average expert redundancy (co-activation) across each layer \textcolor{NavyBlue}{before} and \textcolor{BurntOrange}{after} multi-modal fine-tuning. The observed reduction in average expert redundancy after fine-tuning indicates that redundant experts were leveraged to learn the new modality.}
\label{fig:exp_redundancy}
\end{figure}

We quantify redundancy among experts using the expert co-activation (ECA) metric. ECA is defined as the proportion of instances in which two specific experts are activated simultaneously, normalized by the total number of activations of one of those experts \citep{muennighoff2024olmoe}. High co-activation values indicate that experts frequently activate together, suggesting that they may be functionally redundant. Formally, for two experts \( E_i \) and \( E_j \), with respective activation frequency, $N_{E_i}$ and $N_{E_j}$, ECA can be defined as:
\begin{equation}
    ECA(E_i, E_j) = \frac{N_{E_eE_j}}{N_{E_i}}, 
\end{equation}

where, $N_{E_eE_j}$ denotes the frequency of simultaneous expert activation. 

Fig. \ref{fig:exp_redundancy} illustrates the average ECA across all experts in each layer. Firstly, we observe that experts in the original pre-trained MoE exhibit substantial redundancy across layers. In contrast, after multi-modal fine-tuning, this redundancy is markedly reduced, especially in the initial layers of the model. Overall, this reduction implies that the model has effectively leveraged its inherent redundancy as the latent capacity to learn the new modality, confirming our hypothesis.

\subsubsection{Emergence of Modality-specific and Modality-agnostic Experts} \label{sec:modality_specific_agnostic}

\begin{figure}[ht]
    \centering
    \begin{subfigure}[b]{0.85\textwidth}
        \includegraphics[width=\textwidth]{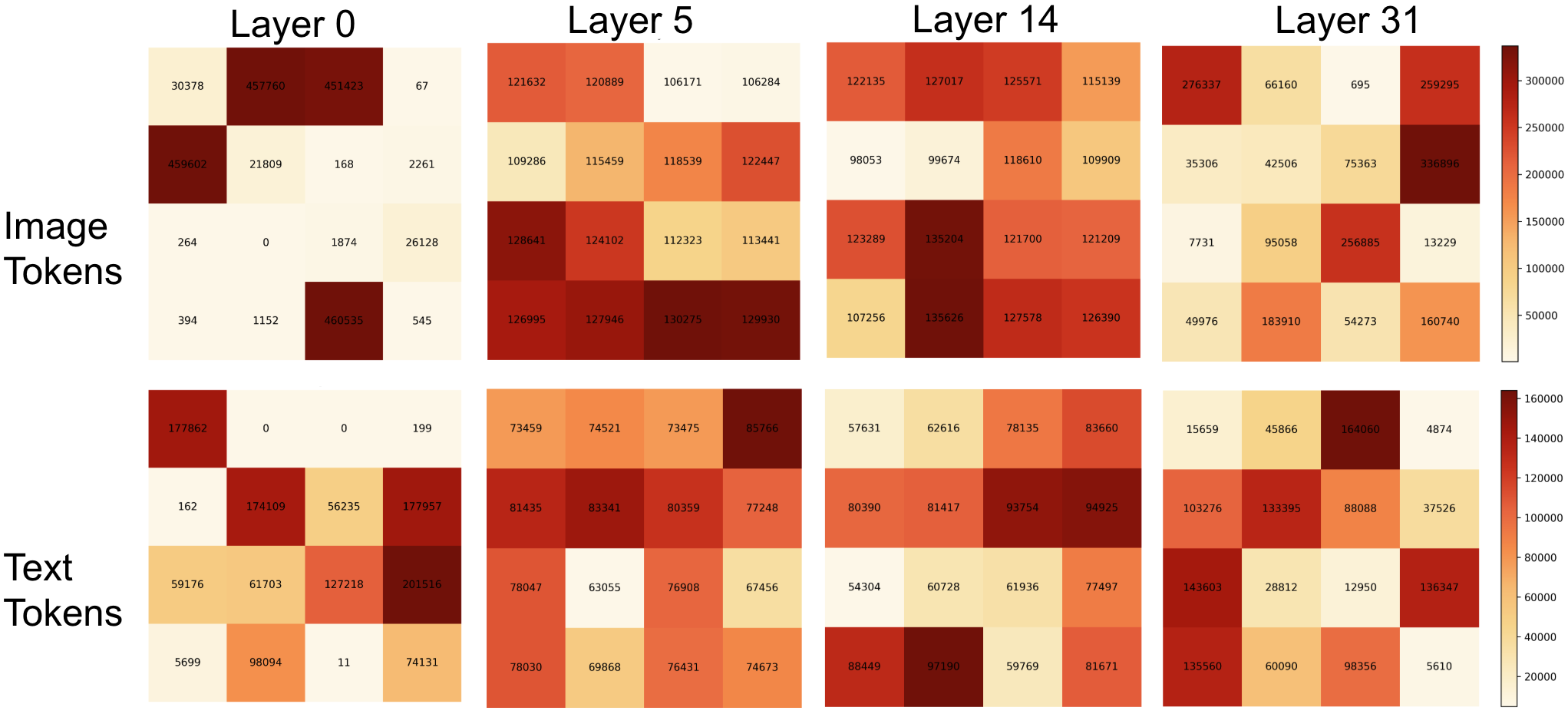}
        \caption{Number of tokens routed (routing preferences) to each of the 16 experts in the first, middle and last layers from image (top) and text modalities (bottom) for \textbf{PLoRA}.}
        \label{fig:modality_specific_vis}
    \end{subfigure}
    
    \vspace{0.5cm} 
    
    \begin{subfigure}[b]{0.85\textwidth}
        \includegraphics[width=0.8\textwidth]{assets/LoRA_FT.pdf}
        \caption{Expert routing preferences in the first layer for image and text tokens in case of conventional \textbf{LoRA} fine-tuning. }
        \label{fig:lora_ft}
    \end{subfigure}
    
    \caption{Contrasting the routing preferences for \textcolor{ForestGreen}{PLoRA} and \textcolor{BurntOrange}{LoRA} fine-tuning. While PLoRA fine-tuning learns to build \textit{modality-exclusive} pathways, LoRA fails to do so, resulting in a performance degradation in the original capabilities.}
    \label{fig:stacked_figures}
\end{figure}


We examine the routing preferences for input tokens across modalities. Specifically, we visualize the frequency with which tokens are assigned to each of the 16 experts in different model layers for both image and text inputs (see Fig. \ref{fig:modality_specific_vis}). This analysis reveals two key phenomena. \textbf{First}, tokens from each modality display pronounced exclusivity in their routing in the early and late model layers. In other words, the experts most frequently chosen by image tokens are amongst the least chosen ones by text tokens and vice versa. For example, in Layer 0, experts 1, 2, 4, and 14 are predominantly selected for image tokens, whereas these experts are among the \textit{least} selected for text tokens, a trend that similarly appears in Layer 31. 
\textbf{Second}, while the middle layers exhibit some degree of modality-specific specialization of experts, this effect is notably less pronounced. 
These observations suggest the desired and expected emergence of \textit{modality-specific} experts in the early and late layers, and \textit{modality-agnostic} semantics-focused experts in the middle layers after performing multi-modal fine-tuning.

\textbf{LoRA Fine-tuning Hinders Expert Exclusivity:} We visualize the routing tendencies with conventional LoRA fine-tuning, highlighting the expert selections in the first layer. In this case, experts 0, 4, 5, 11, and 15 are frequently chosen for both image and text modalities, contrasting with PLoRA, which exhibits high expert \textit{exclusivity} in the same layer. This overlap in routing helps explain the performance degradation on text benchmarks reported in Tab. \ref{tab:nlp_results}. Routing tokens from both modalities to a common set of experts may effectively \quotes{overwrite} the model's original capabilities.


\section{Discussion}
\vspace{-3mm}
In this work, we extend uni-modal, pre-trained LLMs to multi-modal generation without sacrificing language performance, while keeping parameter and training costs modest. We leverage inherent model redundancy by employing Mixture-of-Experts (MoEs) to create distinct pathways for each modality. Furthermore, we introduce novel components such as a Gromov-Wasserstein–based parameter initialization scheme improving convergence and stability and Partial LoRA (PLoRA) to preserve language generation. Our results demonstrate robust image generation with no effective degradation in text performance at minimal computational cost. Extensive analysis reveals the emergence of both modality-specific and modality-agnostic experts, along with reduced overall redundancy. Overall, our framework provides a new approach for transitioning from uni-modal to multi-modal LLMs. 

\textbf{Future Directions:} Our work is an initial step in leveraging model redundancy for new modality learning, with several potential improvements. First, scaling up with larger datasets \citep{schuhmann2022laion},  LLMs featuring expanded vocabularies \citep{grattafiori2024llama,yang2024qwen2} and specialized image tokenizers \citep{ge2023planting,ge2024making} could significantly enhance performance, as suggested by scaling laws \citep{kaplan2020scalinglawsneurallanguage}. Additionally, larger MoE models with more experts \citep{muennighoff2024olmoe} may provide greater latent capacity for multi-modal learning. Second, rather than treating all modalities equally during fine-tuning, a specialized routing mechanism \citep{huang2024harder} could dynamically allocate experts to the more challenging new modality. Finally, shared expert utilization \citep{dai2024deepseekmoe}, which keeps certain experts consistently active, may further boost cross-modal correlations and overall generative performance.

\newpage

\bibliography{colm2025_conference}
\bibliographystyle{colm2025_conference}

\newpage
\appendix
\section{Appendix}

\subsection{Multi-Modal Generation via Unified Architecture} \label{sec:mm_generation_appendix}
\begin{figure}[ht]
\begin{center}
\includegraphics[width=0.85\columnwidth]{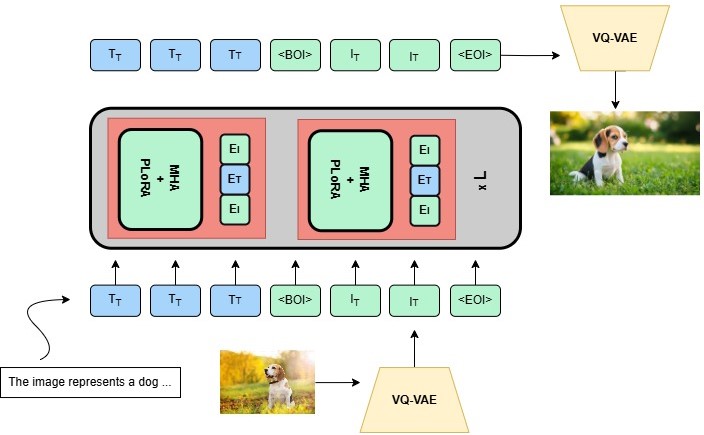}
\end{center}
\caption{Overview of the learning process: The input image is tokenized into discrete tokens using a VQ-VAE encoder and combined with text tokens, separated by special tokens indicating the start and end of the image tokens. The LLM is trained using the next-token prediction objective, and the generated image is reconstructed using the VQ-VAE decoder.}
\label{fig:mm_generation}
\end{figure} 

To achieve a unified architecture for multiple modalities (image and text in our case), the first step is to convert the new input modality into discrete token sequences using modality-specific encoders, which can then be trained with the standard next-token prediction loss employed by LLMs. For images, this tokenization is typically performed using a Vector-Quantized Variational AutoEncoder (VQ-VAE) \citep{van2017neural}. A VQ-VAE transforms the input image pixels $x$ into a corresponding feature map $f$ and assigns each vector $f^{(i,j)}$ in the feature map to the index $q_{(i,j)}$ of its closest codebook vector $z^{(i,j)}$. During decoding, the indices $q_{(i,j)}$ are mapped back to their respective codebook vectors $z^{(i,j)}$ which are then reconstructed into the image pixels $\hat{x}$ by the decoder. 
We employed the image tokenizer in \citet{sun2024autoregressive}. This tokenizer has a codebook size of 16384 i.e. each image is converted into 16384 discrete tokens.

In order for the LLM to interpret these \textit{new} tokens, we proceed as follows. We expand the tokenizer's vocabulary by adding 16384 tokens corresponding to images, along with two special tokens, \textbf{\texttt{<|boi|>}} and \textbf{\texttt{<|eoi|>}}, which indicate the beginning and end of an image in the input sequence, respectively.
To encode and decode these tokens, we further enlarge the embedding and head layers of the LLM. Formally, let the number of new tokens (image tokens plus special tokens) be $T$, let $|V_t|$ denote the size of the text vocabulary, and let $d$ denote the embedding dimension. The original embedding and head layers have shapes $|V_t| \times d$. After incorporating the new tokens, the total parameter count becomes $(|V_t| + T) \times d$, meaning that an additional $T \times d$ parameters have been introduced.
For example, in our implementation using the \quotes{\textit{LLaMA-MoE (4/16)}} model where the embedding dimension is 4096 and the original vocabulary size is 32000, the expansion resulted in the addition of $16386 \times 4096 \approx 67M$ parameters. \textbf{Note that} this expansion is a standard procedure when incorporating a new modality in a unified architecture and \textbf{does not indicate} any parameter inefficiency in our approach.

\textbf{\textcolor{orange}{Note on Initialization Schemes:} } Continuing from the formal notations above, our goal is to initialize the newly added \(T \times d\) parameters in the embedding and head layers so that the pairwise distance distributions of the pre-trained text embeddings (\(|V_t| \times d\)) and the new embeddings (\(T \times d\)) are aligned. To achieve this, we propose a novel initialization scheme based on the Gromov-Wasserstein (GW) distance (see Sec. \ref{sec:gw_init}). For comparison, under \textbf{Random Initialization}, the new parameters are assigned random values, while under \textbf{Mean Initialization}, they are set to a constant equal to the mean of the existing \(|V_t| \times d\) embeddings. During training, we set only the new embeddings as trainable and keep the original embeddings frozen.




\subsection{Contrasting our approach with MARS}

\begin{figure}[ht]
\begin{center}
\includegraphics[width=0.8\columnwidth]{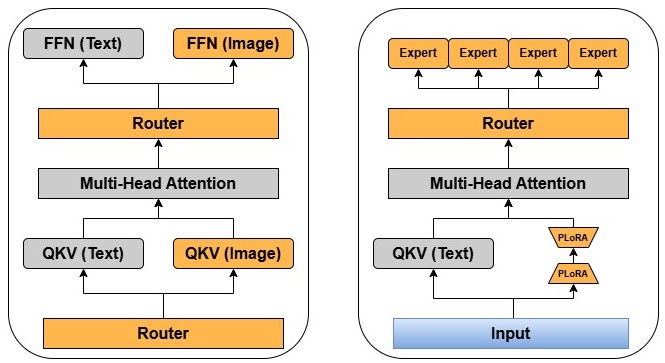}
\end{center}
\vspace{-10pt}
\caption{We contrast MARS~\citep{he2024mars} with our approach. \textbf{Left:} MARS doubles parameters with its SemVIE module by duplicating QKV and FFN modules for visual tokens. \textbf{Right:} Our framework minimally increases parameters using PLoRA, with expert parameter counts matching the original FFN.}
\label{fig:mars_moe}
\end{figure}

We contrast our solution with that of \citet{he2024mars} (MARS) in Fig. \ref{fig:mars_moe} and Tab. \ref{tab:mars_comparison}. MARS allocates distinct modules to handle image and text modalities. Specifically, both the attention and FFN modules (i.e., QKV and FFN, respectively) are duplicated across all model layers, with a routing mechanism placed before these modules to direct modality-specific tokens to separate pathways. During training, the original model parameters remain frozen, while only the new parameters (highlighted in \textcolor{orange}{orange}) are updated. Although effective, this approach more than doubles the original parameter count for learning each modality. \\
In contrast, our approach introduces PLoRA parameters in place of duplicating the entire attention module. We then partition the FFN module into several smaller, equally sized experts and place a router before these experts to establish modality-specific pathways. During training, only the newly introduced parameters (PLoRA, router, and experts) are optimized, while the rest of the model remains frozen. This strategy provides similar capabilities at a significantly lower computational cost.

\subsubsection{Parameter, Data, and Compute Budget}
\begin{table}[ht]
\resizebox{\textwidth}{!}{%
\begin{tabular}{@{}lll@{}}
\toprule
\textbf{}                    & \textbf{MARS}   & \textbf{LLaMA-MoE-PLoRA} \\ \midrule \\
\textbf{Base LLM}            & Qwen-7B \citep{bai2023qwen}        & LLaMA-2-7B \citep{touvron2023llama}              \\
\textbf{Base LLM Type}            & Multi-Modal VLM & Uni-Modal LLM            \\
\textbf{Base LLM Params}     & 7B              & 7B                       \\
\textbf{New Modality Params} & 7B              & 0.0083B                  \\
\textbf{Training Data}       & 250M            & 7.5M                     \\
\textbf{A100 GPU Days}       & 587             & 180 (approx)              \\ \bottomrule
\end{tabular}
}
\caption{A comparison with MARS on Parameter, Data, and Compute Budget.}
\label{tab:mars_comparison}
\end{table}

We further contrast our approach with MARS in Tab. \ref{tab:mars_comparison} in terms of parameter, data and compute budget. 

\textcolor{NavyBlue}{\textbf{Parameter Budget: }}The parameter count of the base LLM employed in both approaches is the same (7B), however, MARS employs a VLM, which already understands the image modality. The number of parameters introduced specifically for learning the new modality is denoted by \quotes{\textbf{New Modality Params}}. In the case of MARS,  the SemVIE module introduces 7B parameters, effectively doubling the parameter count. For LLaMA-MoE, PLoRA parameters account for only 0.008B parameters (875x reduction).

\textcolor{NavyBlue}{\textbf{Data Budget: }} MARS was trained on 250M samples where 200M samples were used for Stage-1 training and an additional 50M high-quality samples were employed for Stage-2 training. On the contrary, our approach employed just 7.5M high-quality samples. 

\textcolor{NavyBlue}{\textbf{Compute Budget: }} MARS conducted training for a total of 587 A100 GPU days as opposed to just 180 A100 GPU days for our approach. Note that we provide an approximation of \quotes{A100 GPU days} since our training was conducted on Nvidia L40 GPUs, which have significantly smaller VRAM (48GB as compared to 80GB in A100s).




\subsection{Examples of Strong Textual Coherence} \label{sec:text_coherence}
\begin{figure}[ht]
\begin{center}
\includegraphics[width=\columnwidth]{assets/textual_coherence.pdf}
\end{center}
\caption{Figure illustrating examples of strong textual coherence in the generated samples. In each case, key details of the prompt that are accurately captured in the generated sample are \underline{\textbf{highlighted}}, underscoring the model's strong ability to adhere to the text prompt.}
\label{fig:text_coherence}
\end{figure}

\newpage
\subsection{More Generated Samples} \label{sec:more_examples}

\begin{figure}[ht]
\begin{center}
\includegraphics[width=\columnwidth]{assets/generated_samples_appendix1.pdf}
\end{center}
\label{fig:generated_samples_appendix1}
\end{figure}

\begin{figure}[ht]
\begin{center}
\includegraphics[width=\columnwidth]{assets/generated_samples_appendix2.pdf}
\end{center}
\label{fig:generated_samples_appendix2}
\end{figure}

\begin{figure}[ht]
\begin{center}
\includegraphics[width=\columnwidth]{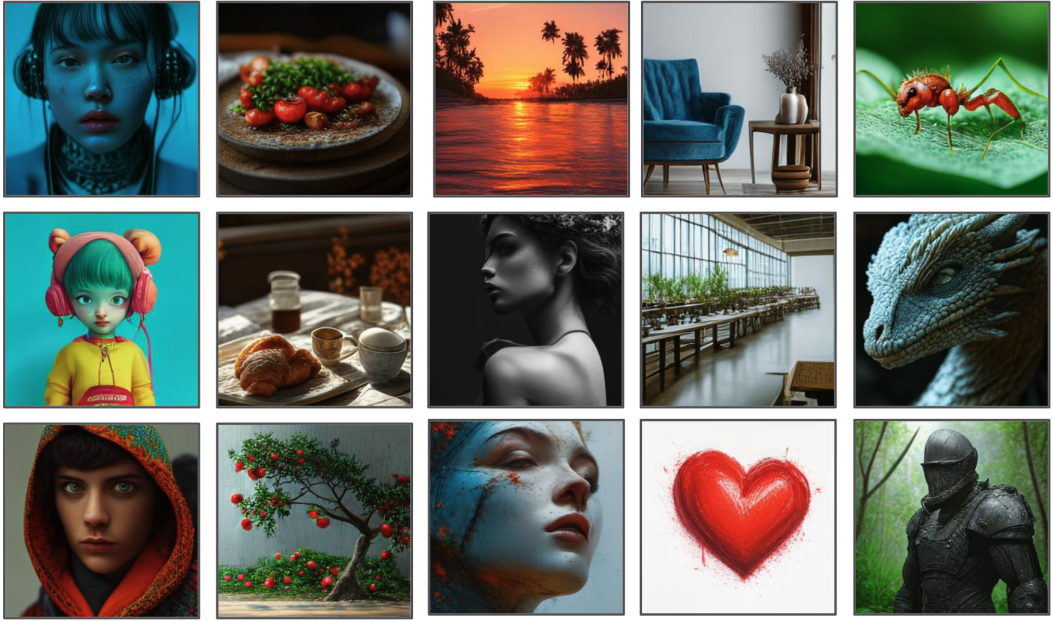}
\end{center}
\caption{More examples of generated samples.}
\label{fig:generated_samples_appendix3}
\end{figure}



\newpage
\subsection{Generation Prompts} \label{sec:gen_prompts}

In this section, we present the text prompts used for generating the images in Fig. \ref{fig:generated_samples}.

\textbf{Top Row | Cols 1-5}
\begin{enumerate}
    \item In the image, there is a young woman who is the main subject. She is adorned in a vibrant red dress that contrasts beautifully with the black wall behind her. The dress features a ruffled neckline and sleeves, adding a touch of elegance to her attire. On her head, she wears a wide-brimmed red hat, which matches her dress and adds a pop of color to the scene. The woman's gaze is directed off to the side, giving her a thoughtful and serious expression. This, combined with her direct gaze into the camera, creates a captivating portrait. Her hair, styled in loose waves, frames her face and complements her overall look. The image does not contain any text or other discernible objects. The focus is solely on the woman, her attire, and her expression. The relative position of the woman to the black wall suggests she is standing quite close to it. The image captures a single moment in time, with no indication of movement or action. It's a still portrait that tells a story through its subject and her attire.

    \item The image captures a close-up of a woman's face, bathed in soft light. Her eyes are gently closed, and her lips are slightly parted, as if she's about to speak. The focus is on her nose and forehead, which are adorned with small droplets of water. The droplets, glistening under the light, add a sense of freshness to her appearance. The background is a dark blue-green color, providing a stark contrast to the woman's skin tone. This contrast accentuates the details of her face, making them stand out even more. The image does not contain any text or other discernible objects. The relative position of the woman's face to the background suggests that she is the main subject of this image. The overall composition of the image is simple yet striking, with the woman's face being the focal point.

    \item In the image, a white and gray cat with striking blue eyes is the main subject. The cat's fur is long and shaggy, giving it a fluffy appearance. Its ears are pointed upwards, adding to its alert and curious expression. The cat is looking to the left of the frame, as if something has caught its attention. The cat is positioned in front of a window, which is blurred in the background, suggesting a depth of field effect from the camera. The window allows light to filter into the room, casting a soft glow on the cat. The cat's gaze and the direction of the light create a sense of interaction between the viewer and the scene. The image captures a quiet moment in the cat's day, providing a glimpse into its world. The colors, lighting, and composition all contribute to a serene and captivating image.

    \item The image presents a close-up view of a woman's face, captured in profile. Her eyes are gently closed, and her lips are slightly parted as if she's about to speak or sing. The woman's face is adorned with a vibrant array of colors and patterns, creating a mosaic-like effect that covers most of her visage. The colors span a wide spectrum, including hues of blue, orange, red, and yellow, which stand out vividly against the stark white of her skin. The patterns on her face are intricate and varied, with geometric shapes and swirls interspersed throughout. These patterns add a dynamic element to the image, making the woman's face appear as if it's telling a story or expressing an emotion. The background of the image is a light beige color, speckled with small black dots scattered randomly across it. This backdrop provides a neutral canvas that allows the colors and patterns on the woman's face to take center stage. Overall, the image is a striking piece of art that uses color, pattern, and perspective to create a captivating visual narrative. The woman's face, with its colorful mosaic-like design, is the focal point of the image, drawing the viewer's attention and inviting them to explore the story behind the artwork.

    \item The image presents a captivating digital art piece featuring a woman's face. The woman's face, which is the central focus of the image, is rendered in a realistic style. Her features are accentuated with a palette dominated by shades of blue and gray, lending an air of tranquility to her expression. Her eyes, painted in a deep shade of blue, gaze upwards and to the left, as if lost in thought or perhaps gazing at something beyond the frame of the image. Her lips, painted in a soft pink hue, add a touch of warmth to the cool color scheme. The background of the image is a stark white, providing a contrast that makes the woman's face stand out. Adding an element of intrigue to the image are black lines and splatters that surround the woman's face. These elements appear to be abstract brushstrokes, further enhancing the digital art style of the piece. Overall, the image is a beautiful blend of color and form, with each element carefully placed to create a harmonious composition. The use of color and form to convey emotion and mood is a testament to the skill and creativity of the artist.
\end{enumerate}

\textbf{Middle Row | Cols 1-5}

\begin{enumerate}
    \item The image presents a close-up view of a young woman's face, captured in a digital art style. Her eyes, a striking shade of blue, are the focal point of the image, radiating a sense of calm and tranquility. Her hair, a vibrant shade of pink, is styled in loose curls that frame her face, adding a touch of whimsy to the overall composition.  She is adorned with a pair of silver earrings, which add a subtle sparkle to her appearance. The background is a blurred mix of pink and white hues, providing a soft contrast that allows the woman's features to stand out. The image does not contain any discernible text or additional objects. The relative position of the woman to the background suggests she is centrally located within the frame. The image does not provide any information about the woman's actions, as it appears to be a still portrait.  This detailed description is based on the visible elements in the image and does not include any speculative or imaginary content.

    \item In the image, a man is captured in a close-up portrait. He is adorned with a red and green knit hat, which is decorated with white pom poms at the top. The hat's vibrant colors contrast beautifully with his black beard and mustache. His gaze is directed straight at the camera, creating a sense of connection with the viewer. The background of the image is blurred, drawing focus to the man. It appears to be a room filled with Christmas lights, adding a festive atmosphere to the scene. The image does not contain any discernible text. The man's position relative to the background suggests he is standing in front of the lights. The overall composition of the image places the man as the central focus, with the Christmas lights serving as a secondary element in the background.

    \item In the image, a turtle is the main subject, captured in a close-up shot. The turtle 's shell is a striking pattern of black and orange, adorned with intricate designs that add to its charm. The turtle is situated on a bed of small rocks, which provide a contrasting texture to the smoothness of the turtle's shell. The rocks are scattered around the turtle, some closer to the camera and others further away, creating a sense of depth in the image. The background is a soft blur of green foliage, providing a natural backdrop that allows the turtle to stand out. The sun is shining brightly in the top left corner of the image, casting a warm glow over the scene and creating a lens flare that adds a touch of magic to the image.  Overall, the image captures a serene moment in nature, with the turtle as the star of the scene. The turtle's vibrant colors, the detailed patterns on its shell, and the tranquil setting all combine to create a captivating image

    \item In the image, a young girl with curly hair and glasses is the central figure. She is dressed in a white blouse and a blue apron, adding a touch of charm to her appearance. In her hands, she holds two distinct objects - a black cat and a yellow orb. The cat, with its fur as dark as night, is comfortably perched on her shoulder, while the orb, glowing with a warm yellow light, is held in her other hand. The setting appears to be a cozy room, filled with various objects that give it a lived-in feel. A bookshelf filled with books suggests a love for literature, while a clock on the wall indicates the passage of time. A plant adds a touch of greenery to the room, creating a harmonious blend of indoor and outdoor elements. The precise locations of these objects create a well-balanced composition, with the girl and her cat at the center, drawing the viewer's attention. The image does not contain any discernible text. The relative positions of the objects suggest a quiet, peaceful moment captured in time. The girl, the cat, and the orb are all in close proximity, suggesting a bond between them. The room serves as a backdrop, framing the scene and adding depth to the image.

    \item In the image, a woman is captured in a close-up shot, her face adorned with intricate makeup and a traditional Native American headdress. The headdress, a striking feature, is decorated with feathers in hues of brown, red, and white. The woman's gaze is directed towards the left side of the frame, her eyes accentuated by long, dark lashes.  In her hand, she holds a pipe, a symbol often associated with Native Americans. The background of the image is blurred, drawing focus to the woman. However, it's discernible that the setting is outdoors, possibly a forest or a field, adding a natural element to the composition.  The image does not contain any discernible text. The relative positions of the objects suggest a sense of depth, with the woman in the foreground and the forest or field in the background. The woman, the pipe, and the headdress are the main elements in the image, while the background provides context to the setting. The image does not provide any information that allows for a confident count of the objects in the background.  Overall, the image captures a moment of stillness, with the woman in her traditional attire, the pipe in her hand, and the natural backdrop. The precise locations of the objects cannot be determined from the image alone. The image does not contain any imaginary content; everything described can be confidently determined from the image itself.

\end{enumerate}

\textbf{Bottom Row | Cols 1-5}

\begin{enumerate}
    \item The image captures a close-up of a man's face, his features etched with the lines of age and experience. His eyes, a striking shade of blue, gaze directly into the camera, unflinching and intense. His nose, prominent and well-defined, stands out against the rest of his face. The skin of his face is weathered and wrinkled, a testament to the years he has lived. He is dressed in a black jacket, its dark color contrasting with the lighter tones of his face. The background is a blurred gray, a neutral backdrop that further emphasizes the man's face. The image does not contain any discernible text or other objects. The man's position relative to the camera and the background suggests he is the main subject of this image. The image does not provide any information about the man's actions, as he appears to be in a state of stillness. The image is devoid of any aesthetic descriptions, focusing solely on the man and his immediate surroundings.

    \item The image portrays a young boy, adorned in regal attire, exuding an air of majesty and nobility. His gaze is directed straight at the camera, his expression serious, perhaps reflecting the solemnity of his attire. His head is crowned with a gold crown, which is intricately designed with a cross at its center, symbolizing authority and power. Complementing the crown, he wears a gold necklace around his neck, adding to his royal appearance. Over his shoulders, he drapes a large, ornate cape. The cape is richly decorated with gold embroidery, showcasing the meticulous craftsmanship involved in its creation. The fabric of the cape appears to be of a luxurious nature, enhancing the overall grandeur of the boy's attire. The background of the image features a green curtain, providing a stark contrast to the boy's golden attire. The curtain's texture and color add depth to the image, framing the boy and drawing attention to him as the focal point. Overall, the image captures a moment of quiet dignity and regal elegance, embodied by the young boy in his ornate attire. The precise positioning of the objects and the boy's direct gaze create a sense of engagement with the viewer, inviting them to appreciate the intricate details and the overall composition of the image.

    \item In the image, a small dog with a coat of brown and white fur is the main subject. The dog's eyes, a striking shade of blue, are gazing directly into the camera, giving it a curious and alert expression. Adding to its charm is a pink collar around its neck, from which hangs a silver tag.  The dog is not just any ordinary pet, it's a service dog, as indicated by the black harness it's wearing. The harness is equipped with a silver buckle, matching the silver tag on its collar.  The setting of the image is equally captivating. The dog is standing on a road that appears to be at sunset. The sky, painted in hues of orange and yellow, suggests that the sun is setting, casting a warm glow over the scene. In the distance, you can see trees standing tall, their silhouettes adding depth to the landscape. Overall, the image beautifully captures a moment in the life of this service dog, set against the backdrop of a serene sunset.

    \item The image presents a captivating scene of a fox, rendered in vibrant hues of orange and brown, with striking red accents on its face and ears. The fox is captured in profile, its gaze directed towards the right side of the image, as if gazing into the distance. It stands amidst a lush array of green foliage and flowers, adding a touch of nature's charm to the composition. The entire scene is encapsulated within a circular frame, lending a sense of completeness to the image. The art style is reminiscent of stained glass, with the fox and the surrounding flora intricately intertwined, creating a harmonious blend of colors and shapes. The image does not contain any discernible text or countable objects, and there are no explicit actions taking place. The relative positions of the objects suggest a serene coexistence, with the fox and the flora existing in harmony within their shared space. The image is a testament to the beauty of nature, captured in a moment of tranquility.

    \item In the image, a figure clad in a futuristic suit of green and gray is seated in a chair. The suit is detailed with a chest plate. The figure's head is protected by a helmet equipped with a visor that exhibits a striking red and orange glow.  The figure appears to be engrossed in reading a piece of paper that rests on their lap. The setting is a dimly lit room. The overall atmosphere of the image suggests a scene straight out of a science fiction narrative.
\end{enumerate}

\subsection{MoE with Fine-Grained Expert Segmentation}
We hypothesize that learning across multiple modalities requires the experts to (1) capture more fine-grained details in the data and (2) exhibit greater flexibility to enable better adaptive combinations of activated experts. \\
During training, each expert in the model specializes in learning distinct aspects of the input, driven by the diverse tokens it processes. As input tokens are routed to specific experts, each one focuses on capturing unique nuances or features. However, when the number of experts is relatively small, each expert is tasked with learning a wide range of information. This leads to a challenge where the broad knowledge acquired by an expert cannot be efficiently utilized simultaneously, potentially limiting the model’s performance. \\
To address this challenge, we adopt the principle of \textbf{\textit{Fine-Grained Expert Segmentation}} \citep{dai2024deepseekmoe}. 
Each expert is subdivided into smaller, more specialized units, effectively increasing the total number of experts while maintaining the overall parameter activation constant. This segmentation allows each expert to focus on learning finer details of the data, enhancing the model’s adaptability and improving its ability to combine the activated experts effectively. Our expert construction method (Sec. \ref{sec:llm_moe}) is designed based on this principle.

\subsection{Visualizing Modality-Specific Routing Preferences}

\begin{figure}[ht]
\begin{center}
\includegraphics[width=\columnwidth]{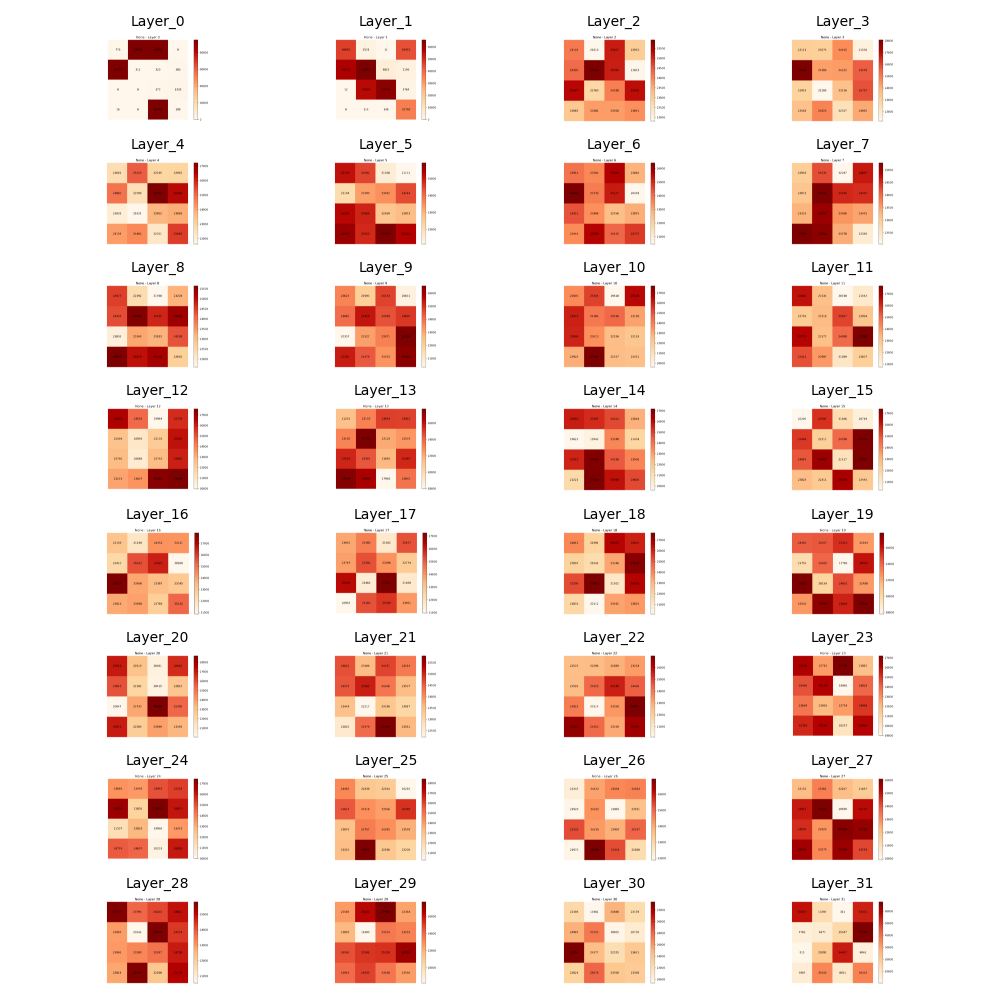}
\end{center}
\caption{Figure representing the routing preferences for \textit{image tokens} across each layer of the model.}
\label{fig:img_modality_specific}
\end{figure}

\begin{figure}[ht]
\begin{center}
\includegraphics[width=\columnwidth]{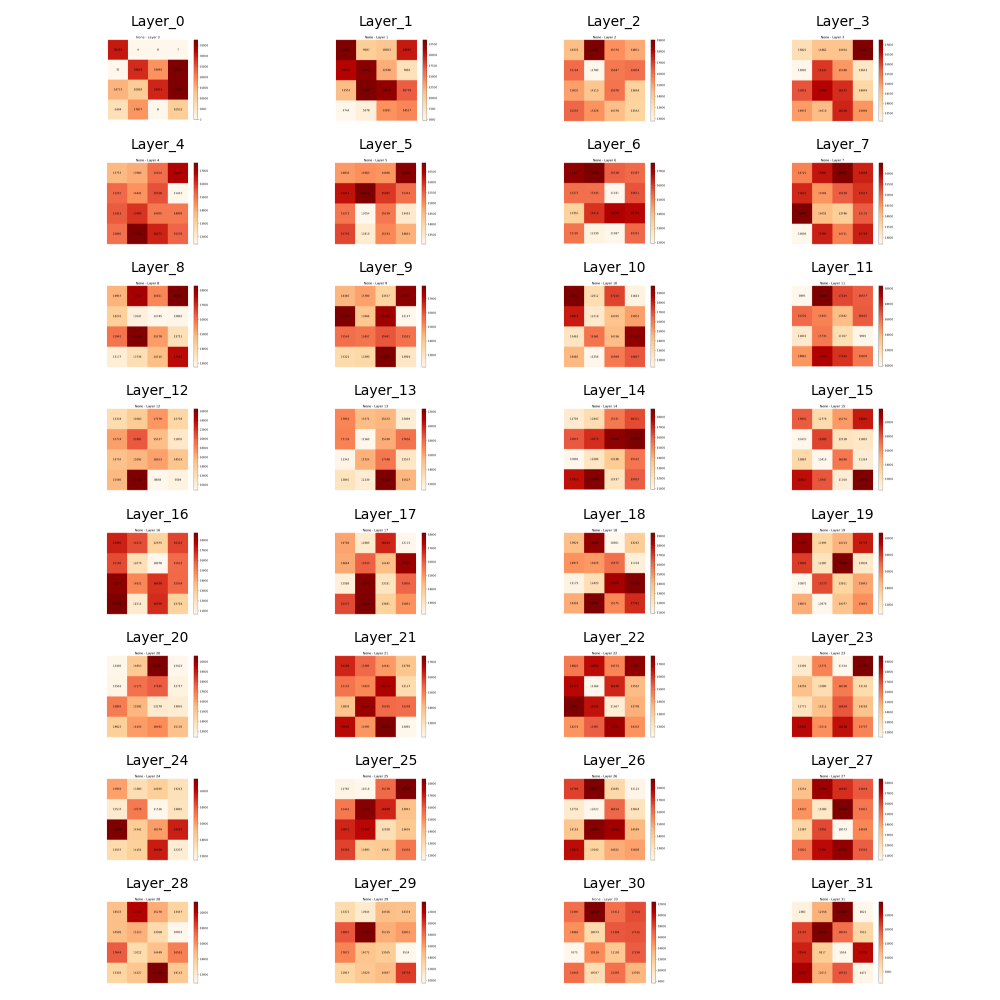}
\end{center}
\caption{Figure representing the routing preferences for \textit{text tokens} across each layer of the model.}
\label{fig:text_modality_specific}
\end{figure}

In this section, we visualize the routing preferences for image and text tokens (Figures \ref{fig:img_modality_specific} and \ref{fig:text_modality_specific} respectively). As demonstrated in section \ref{sec:modality_specific_agnostic}, routing preferences show high specificity for each modality in the initial and final layers of the model, indicating the emergence of \textit{modality-specific} experts. On the contrary, the middle layers demonstrate significantly lower exclusivity indicating the formation of \textit{modality-agnostic} experts.

\end{document}